\def\year{2020}\relax
\documentclass[letterpaper]{article} 
\usepackage{aaai20}  
\usepackage{times}  
\usepackage{helvet} 
\usepackage{courier}  
\usepackage[hyphens]{url}  
\usepackage{graphicx} 
\usepackage{graphicx}  
\nocopyright
\usepackage{subcaption}
\usepackage{amsmath}
\usepackage{booktabs}
\newcommand{\blap}[1]{{\begin{tabular}[t]{@{}l@{}}#1\end{tabular}}}

\title{Accelerating Training using Tensor Decomposition}
\author{Mostafa Elhoushi, Ye Henry Tian, Zihao Chen, Farhan Shafiq, Joey Yiwei Li \\
Toronto Hetrogeneous Compilers Lab \\
19 Allstate Parkway \\
Markham, Ontario, L3R 5B4 \\
Canada
}
\begin{document}

\maketitle

\begin{abstract}
Tensor decomposition is one of the well-known approaches to reduce the latency time and number of parameters of a pre-trained convolutional neural network (CNN) model. However, in this paper, we propose an approach to use tensor decomposition to reduce training time of training a model from scratch. In our approach, we train the model from scratch (i.e., randomly initialized weights) with its original architecture for a small number of epochs, then the model is decomposed, and then continue training the decomposed model till the end. There is an optional step in our approach to convert the decomposed architecture back to the original architecture. We present results of using this approach on both CIFAR10 and Imagenet datasets, and show that there can be upto 2$\times$ speed up in training time with accuracy drop of upto 1.5\% only, and in other cases no accuracy drop.. This training acceleration approach is independent of hardware and is expected to have similar speed ups on both CPU and GPU platforms. The code is submitted along with the paper and will be open-sourced upon acceptance of the paper.
\end{abstract}

\section{Introduction}
While deep learning has obtained high accuracy in computer vision, natural language understanding, and many other fields, one of its main challenges is the extensive computation cost required to train its models. A typical neural network architecture may take upto 2 weeks to train on the Imagenet dataset. Using distributed training, a ResNet50  model can train in 6 minutes, but at the financial and energy cost of 1024 GPUs \cite{Jia2018HighlySD}. The energy required to train the average deep learning model is equivalent to using fossil fuel releasing around 78,000 pounds of carbon, which is more than half of a car's output during its lifetime \cite{Strubell2019EnergyAP}. The financial and time costs of training deep learning models is making it increasingly difficult for small- and medium- sized companies and research labs to explore various architectures and hyperparameters.

While there has been extensive development to meet the increasing computation cost of training by improving hardware design of GPUs - as well as other forms of specialized hardware - and advancements in distributed training using large number of servers and GPUs, there has been less advancement in reducing the computation costs of training. \cite{DBLP:journals/corr/abs-1907-10597} estimated that from 2012 to 2018 the computations required for deep learning research have estimated 300,000 $\times$. 

In this paper we propose a hardware independent method to reduce the computation cost of training using tensor decomposition. A lot of research has been made on compressing pre-trained models using tensor decomposition. However, to the best of our knowledge, this paper is the first to propose to use tensor decomposition during training to reduce the computation cost and training time.

In this paper, we will first present related work in reducing the computation cost and latency time of models during inference, followed by related work in reducing training time. Then, we will explain tensor decomposition, and one of its specific methods - Tucker decomposition - that we use in our solution. We then present our proposed solution to decompose the model during training, followed by the results and performances on CIFAR10 and Imagenet datasets. 

\section{Related Work}
\subsection{Inference Acceleration}
This section covers related works in reducing CNN model size and speeding up training and/or inference which fall in several categories. First, is searching or designing architectures that have lower number of parameters and hence reduce computation latency time while maintaining reasonable prediction accuracy. Proposed models under this category are SqueezeNet \cite{Iandola2017}, MobileNets \cite{Howard2017}, and MobileNetV2 \cite{Sandler2018}. Such methods usually result in considerable drop in accuracy: SqueezeNets have Top1/Top5 accuracies of 58.1\%/80.4\%, MobileNets have 69.6\%/89.1\%, and MobileNets have 71.8\%/90.4\%.

The second category, is replacing a portion or all of convolution operations in a model with operations that require less computation time or less parameters. Binarized Neural Networks \cite{Courbariaux2016}, XNOR-Net \cite{Rastegari2016}, Bi-Real Net \cite{Liu2018} and ABCNet \cite{Lin2017BCNN} replaced multiplication in convolution operation with logical XNOR operation. Such XNOR-based models usually use regular convolutions in training and therefore do not speed up training, although they speed up inference and reduce parameter size. \cite{Chen2019DropAO} proposed Octave Convolution, a spatialy less-redundant variant of regular convolution. \cite{Wu2017ShiftAZ} proposed pixel-wise shift operation to replace a portion of convolution filters. 

The third category is to quantize the parameters of regular convolution operations from 32-bit floating point presentation to smaller number of bits, such as 8-bit integers. \cite{stock2019bit} used vector-quantization method that is efficient in compressing ResNet models by upto 20x. However, most quantization methods require training using regular convolution till the end, before quantizing, and therefore do not speed up training, although they speed up inference.

The fourth category is pruning: removing a portion of convolution filters in each layer - usually - based on some heurestic. \cite{Hassibi1993} and \cite{LeCun1990} published some of the other earlier works on pruning. \cite{Hur2019} proposed using information entropy of filters as the criteria to prune filters. \cite{DBLP:journals/corr/abs-1905-05212} proposed an end-to-end pruning method using trainable masks to decide which convolution filters should be pruned. \cite{Molchanov_2019_CVPR} used Taylor expansion to estimate the contribution of each filter, and hence prune the filters that contribute less to the model accuracy. While most literature on pruning show that training a pruned model from scratch has lower accuracy then pruning a pretrained model to the same pruned architecture, some researchers have reported otherwise \cite{liu2018rethinking}. Nevertheless, although training a pruned architecture is usually faster than training the original model, obtaining the pruned architecture in the first place requires training the full model in the first place till the end, while our proposed solution only requires training the first 10 or 30 epochs before compressing.

The fifth categroy is tensor decomposition: separating a regular convolution operation into multiple smaller convolutions whose combined number of parameters or combined latency is less than that of the original operation. This will be explained in the next subsection.

\subsection{Tensor Decomposition}
\begin{figure}[t]
\centering
\begin{subfigure}[b]{0.9\columnwidth}
	\centering
	\includegraphics[width=\textwidth]{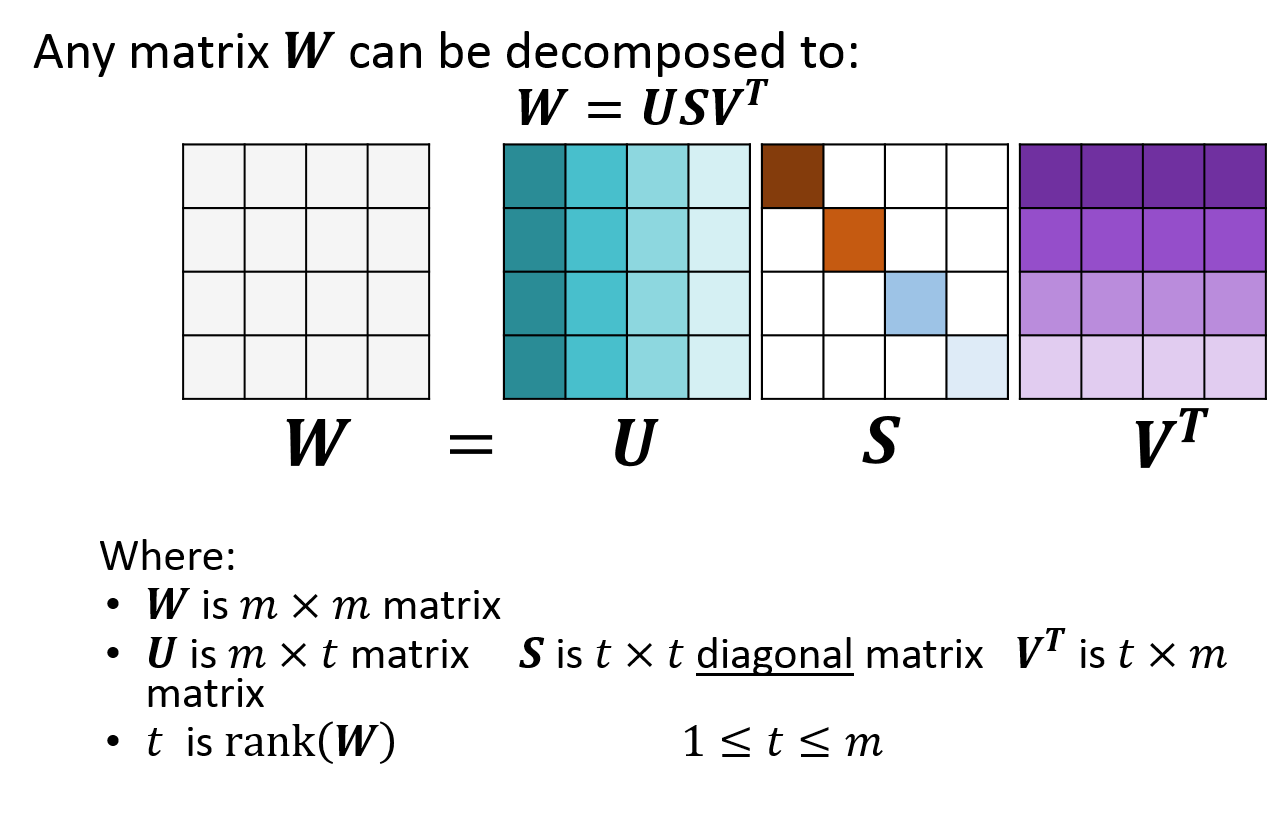}
	\caption{Exact Decomposition}
	\label{fig:svd_exact}
\end{subfigure}
\hfill
\begin{subfigure}[b]{0.9\columnwidth}
	\centering
	\includegraphics[width=\textwidth]{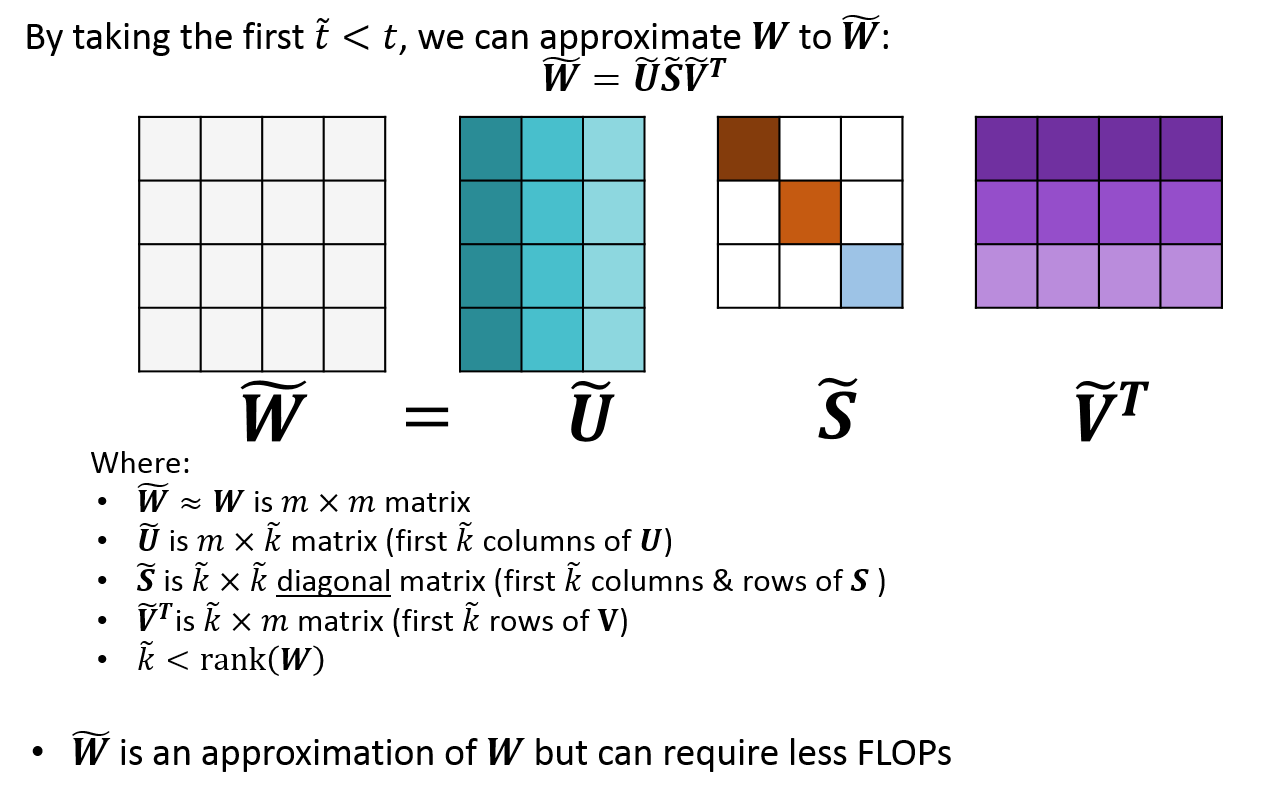}
	\caption{Approximate Decomposition}
	\label{fig:svd_approximation}
	\end{subfigure}
\caption{Singular Value Decomposition.} 
\label{fig:svd}
\end{figure}

\begin{figure}
	\centering
	\begin{subfigure}[b]{0.9\columnwidth}
		\centering
		\includegraphics[width=\textwidth]{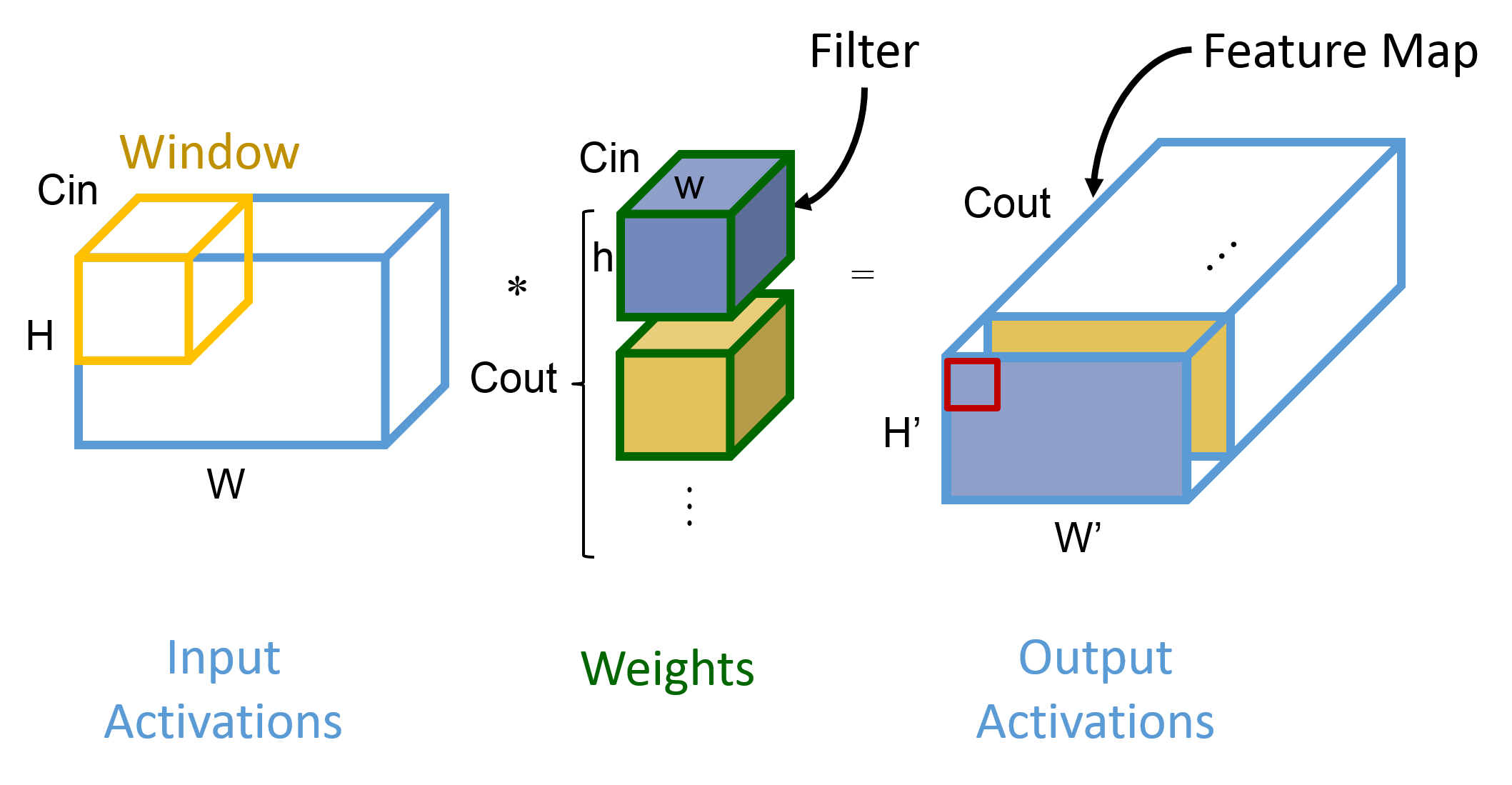}
		\caption{Regular Convolution}
		\label{fig:regular_conv}
	\end{subfigure}
	\hfill
	\begin{subfigure}[b]{0.9\columnwidth}
		\centering
		\includegraphics[width=\textwidth]{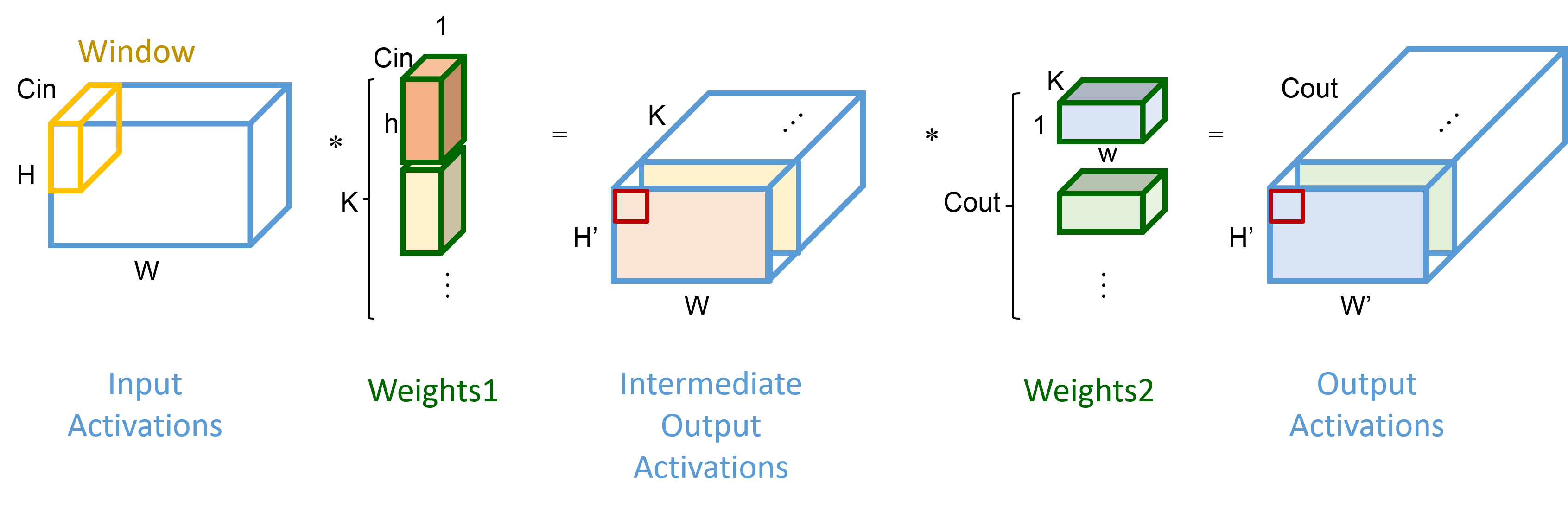}
		\caption{Spatial Decomposition}
		\label{fig:spatial_decomposition}
	\end{subfigure}
	\begin{subfigure}[b]{0.9\columnwidth}
	\centering
	\includegraphics[width=\textwidth]{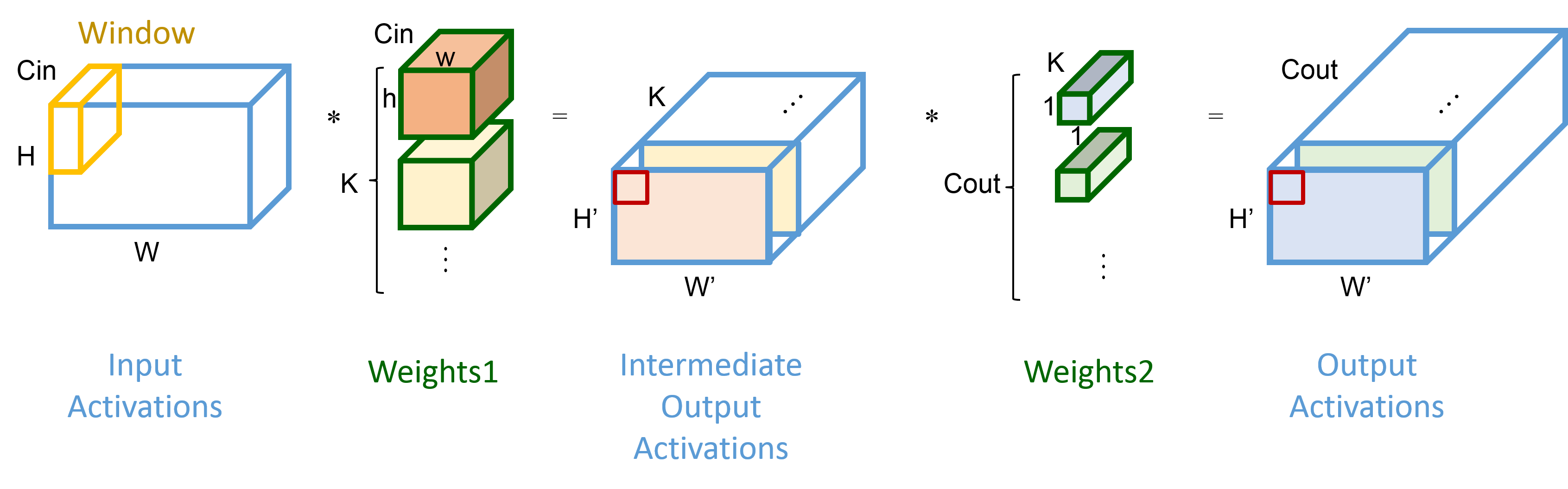}
	\caption{Channel Decomposition}
	\label{fig:channel_decomposition}
	\end{subfigure}
	\begin{subfigure}[b]{0.9\columnwidth}
	\centering
	\includegraphics[width=\textwidth]{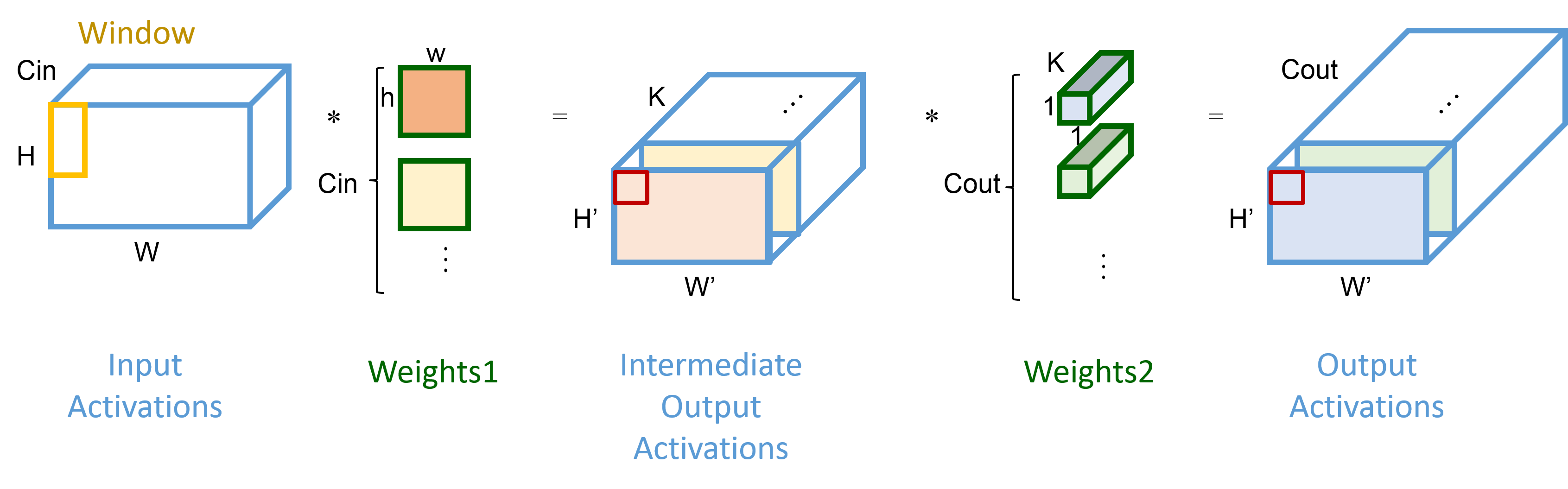}
	\caption{Depthwise Decomposition}
	\label{fig:depthwise_decomposition}
	\end{subfigure}
	\begin{subfigure}[b]{0.9\columnwidth}
	\centering
	\includegraphics[width=\textwidth]{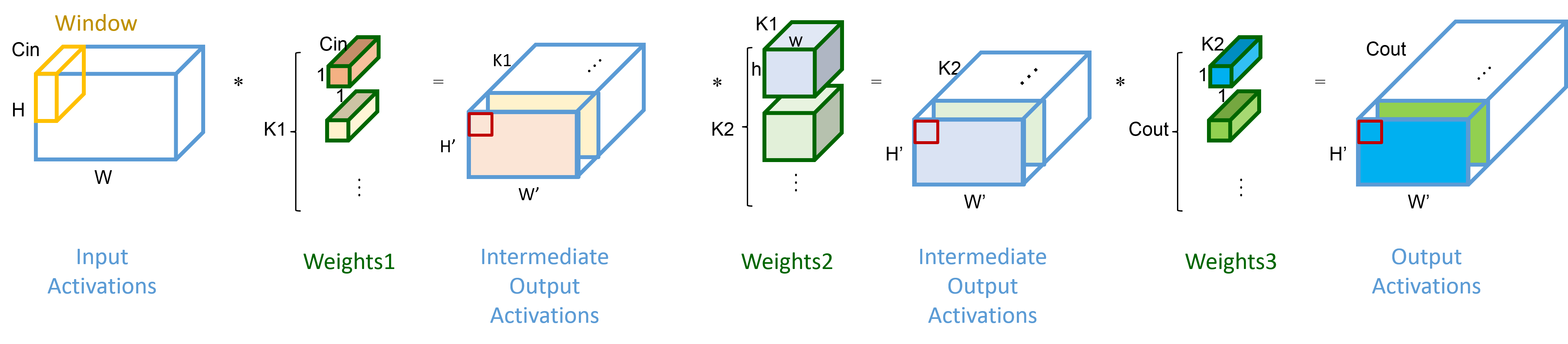}
	\caption{Tucker Decomposition}
	\label{fig:tucker_decomposition}
	\end{subfigure}
	\caption{Illustration of different types of decompositions. Note that $K$, $K_1$, or $K_2$ are always chosen to be less than $C_{out}$ in order to achieve a reduction in the number of parameters.} 
	\label{fig:tensor_decomposition_types}
\end{figure}

Tensor decomposition is based on a concept of linear algebra known as Singular Value Decomposition (SVD) that states that any matrix, $W$, whose dimensions are - without loss of generality - $m \times m$ can be expressed as:
\begin{equation}
W=USV^T
\end{equation}
As shown in Figure \ref{fig:svd}, the dimensions of $U$, $S$, and $V$ are $m \times k$, $k \times k$, $m \times k$. $S$ is referred to as the unitary matrix and it is a diagonal matrix. Each value along the diagonal represent the ``importance'' of the corresponding column of $U$ and row of $V$. The SVD algorithm specifies how to calculate the values of $U$, $S$, $V$ matrices in order to hold this equality. $k$ is known as the rank of the matrix $W$. Most of the time, the rank of a $m \times m$ matrix is $K=m$. If one or more rows and columns of the matrix are linearly dependent on other rows and matrices in the matrix, then the rank is $k<m$. However, in the context of neural networks where the values of the weight matrices are updated during training, this condition is unlikely to happen.

In order to decompose $W$ into terms that have fewer number of parameters, we need to set the rank of the transformation $\tilde{k}$ to be less than the rank, $k$ of the matrix $W$:
\begin{equation}
W \approx \tilde{U}\tilde{S}\tilde{V}^T
\end{equation}
This results in an approximation. In the extreme case of choosing the rank of decomposition $\tilde{k}=1$, the matrix can be represented as $m \times 1 + 1 \times m = 2m$ compared to $m \times m = m^2$ parameters of the original $W$ matrix. For large values of $m$, $2m << m^2$ and hence the storage of the decomposed representation and FLOPs of matrix operation on the decomposed representation is much lower.

The rank has to be selected in an optimal manner in order to balance between the approximation error introduced and the compression obtained. Different methods of rank selection are explained in the following sub-sub-section.

In the context of neural networks, tensor decomposition extends SVD to the 4-dimensional matrices of weights of convolution operators that have dimensions: $C_{out} \times C_{in} \times h \times w$, where $C_{out}$ is the number of channels of the output of the operator, $C_{in}$ is the number of channels of the input image, $h$ is the height of each filter, and $w$ is the width of each filter. There are various types of tensor decompositions as illustrated in Figure \ref{fig:tensor_decomposition_types}: spatial decomposition \cite{8478366}, channel decomposition \cite{Zhang_2016}, depthwise decomposition \cite{DBLP:conf/bmvc/GuoLLCL18}, Tucker decomposition \cite{Tucker1966}. The mathematical expression of each decomposition type and the their derivations from SVD is out of the scope of this paper but they can be found in the reference of each decomposition method. This paper uses Tucker decomposition, and will be explained in more detail in Section \ref{sec:proposed-method}.

\subsubsection{Rank Selection}
Some researchers have used time-consuming trial-and-error to select the optimal rank of decomposition of each convolution layer in a network, by analyzing the final accuracy of the model. \cite{Denton:2014:ELS:2968826.2968968} used alternate least squares. \cite{MacKay91bayesianinterpolation} proposed a data-driven one-shot decision using \textit{empirical Bayes}. In this paper, we used variational Bayesian matrix factorization (VBMF) \cite{NIPS2012_4675}.

\subsection{Training Acceleration}
To speed up training neural networks, the main research efforts in industry gear towards designing and enhancing hardware platforms. Vector units in CPUs enable it to perform arithmetic operations on a group of btyes in parallel. However, CPUs are only practical to train small datasets such as MNIST. NVIDIA's Graphical Processing Units (GPUs) are the most popular hardware platforms used for training. The core principle of GPUs is the existence of hundreds or thousands of cores that perform computation on data in parallel. Other hardware platforms, such as Google’s Tensor Processor Unit (TPU) and Huawei's Neural Processing Unit (NPU), depend on large dedicated matrix multipliers. Other accelerated systems based on custom FPGA and ASIC designs are also explored. 

Another main method to accelerate training is distributed training: training over multiple GPUs on the same workstation, or training over multiple workstations, with each workstation having one or more GPUs. Distributed training methods can be classified into model parallelism methods as well as data parallelism methods. Frameworks and libraries have provided support for distributed training \cite{sergeev2018horovod} to automate the process of distributing workload and accumulating results. A detailed survey of parallel and distributed deep learning is presented in \cite{Ben-Nun:2019:DPD:3359984.3320060}. A key area for research regarding distributed training is to optimize communication between the various workstations \cite{lin2018deep}.

Our proposed method in accelerating training is hardware independent, i.e., does not require specific hardware design. It can build upon the speed ups by faster hardware designs and distributed approaches.

Other hardware independent approaches in literature include \cite{sun17meprop}that  presented a method to speed up training by only updating a portion of the weights during each backpropagation pass. However, the results are only shown for the basic MNIST dataset. \cite{gusmo2016fast} presented an approach to accelerate training by starting with downsampled kernels and input images to a certain number of epochs, before upscaling to the original input image size and kernel size. 

\section{Proposed Method}\label{sec:proposed-method}
We use the end-to-end tensor decomposition scheme similar to that proposed by \cite{DBLP:journals/corr/KimPYCYS15} that in turn uses the Tucker decomposition algorithm proposed by \cite{Tucker1966} and the the rank determined by a global analytic solution of variational Bayesian matrix factorization (VBMF) \cite{NIPS2012_4675}:
\begin{enumerate}
	\item \textbf{Initialize Model}: We start with a model architecture from scratch (i.e., initialized with random weights). 
	\item \textbf{Initial Training}: We then train the weights of the model for a certain number of epochs, e.g., 10 epochs.
	\item \textbf{Decompose}: Then, we decompose the model and its weights using Tucker decomposition and VBMF rank selection.
	The decomposed model has a smaller number of weights then the original model, and hence lower training (and inference time), and a sudden drop in accuracy is expected at that point.
	\item \textbf{Continue Training}: Then, we continue updating those decomposed weights till the end of training.
	\item \textbf{Reconstruction} [Optional]: Before the end of training by a certain number of epochs (e.g., 10) we reconstruct the original architecture by combining the weight matrices of each set of decomposed convolution operations.
	The accuracy at this point does not change, as this reconstruction step is lossless.
	\item \textbf{Fine Tuning} [Optional]: Train the reconstructed model for a few more epochs. 
\end{enumerate}

Tucker decomposition is illustrated in Figure \ref{fig:tucker_decomposition}. To explain Tucker decomposition, we first express a regular convolution operation of weight $\boldsymbol{W}$ with dimensions $C_{out} \times C_{in} \times h \times w$, acting on an input tensor $\boldsymbol{I}$ with dimensions $H \times W \times C_{in} $ to produce an output tensor $\boldsymbol{O}$ with dimensions $H' \times W' \times C_{out}$:
\begin{equation}
\boldsymbol{O}_{x',y',c_{out}}=\sum_{i=1}^{h}\sum_{j=1}^{w}\sum_{c=1}^{C_{in}}\boldsymbol{W}_{i,j,c_{in},c_{out}}I_{x_i,y_j,c_{in}}
\end{equation}
where:
\begin{align}
x_{i} = (x'-1)S + i - P \\
y_{j} = (j'-1)S + j - P \\
\end{align}
where $S$ is the stride and $P$ is the padding.

Tucker decomposition converts this convolution operation into 3 consecutive convolutions:
\begin{align}
\boldsymbol{O^{(1)}}_{x,y,k_{1}}&=&\sum_{c=1}^{C_{in}}\boldsymbol{W^{(1)}}_{c,k_{1}}\boldsymbol{I}_{x,y,c} \\
\boldsymbol{O^{(2)}}_{x',y',k_{2}}&=&\sum_{i=1}^{h}\sum_{j=1}^{w}\sum_{k_1=1}^{K_{1}}\boldsymbol{W^{(2)}}_{i,j,k_{1},k_{2}}\boldsymbol{O^{(1)}}_{x_i,y_j,k_{1}}\\
\boldsymbol{O}_{x',y',c_{out}}&=&\sum_{k_2=1}^{K_{2}}\boldsymbol{W^{(3)}}_{c_{out},k_{2}}\boldsymbol{O^{(2)}}_{x',y',k_{2}}
\end{align}

The first and third convolutions are pointwise convolutions, while the second convolution is a regular spatial convolution with input channels and output channels reduced to $K_1$ and $K_2$ respectively. The Tucker decomposition method described in \cite{Tucker1966} derives the equations to deduce the weights $W^{(1)}$, $W^{(2)}$, $W^{(3)}$ from $W$. The compression ratio of the decomposition is expressed as:
\begin{equation}\label{eqn:compression}
M=\frac{hwC_{in}C_{out}}{C_{in}K_1 + hwK_1K_2 + C_{out}K_2}
\end{equation}
and
\begin{equation}\label{eqn:speedup}
E=\frac{hwC_{in}C_{out}H'W'}{C_{in}K_1HW + hwK_1K_2H'W' + C_{out}K_2H'W'}
\end{equation}
Due to incorporating the height and width of the input and output tensors into the numerator and denominator of the speedup equation, we will notice that speedup in training time is lower than compression ratio.

The values of $K_1$ and $K_2$ are determined by the rank selection method, which is in our case is VBMF. It is out of scope to explain the method here.

It is noteworthy that unlike other compression methods such as pruning and distillation, tensor decomposition can be reversed to retain the original architecture without a change in accuracy in a straightforward manner: by simply performing matrix multiplication of the decomposed matrices. The last 2 steps in the process are only  to show that the overall training process can retain the original, in case if someone would like to use the original architecture - due to some reason - rather than the decomposed smaller architecture. 



\section{Experiments}
We have tested our approach by training VGG19, DenseNet40, ResNet56 on CIFAR10 dataset, and ResNet50 in Imagenet dataset. When training on CIFAR10, the batch size used was 128, and the learning rate was initialized at 0.1, reduced to 0.01 at the 100th epoch, and to 0.001 at the 150th epoch. When training on Imagenet, the batch size was 256 and the learning rate was set to 0.1. 

In addition to training from scratch the original model, as well as training the decomposed model at an early epoch, and the reconstructed model at a late epoch, we also decomposed the original model after it completed training, and fine-tuned for an additional number of epochs. We did this to compare the accuracy and number of parameters of a model decomposed early during training versus a model decomposed after it completed training. 

\begin{figure}[t]
	\centering
	\includegraphics[width=0.9\columnwidth]{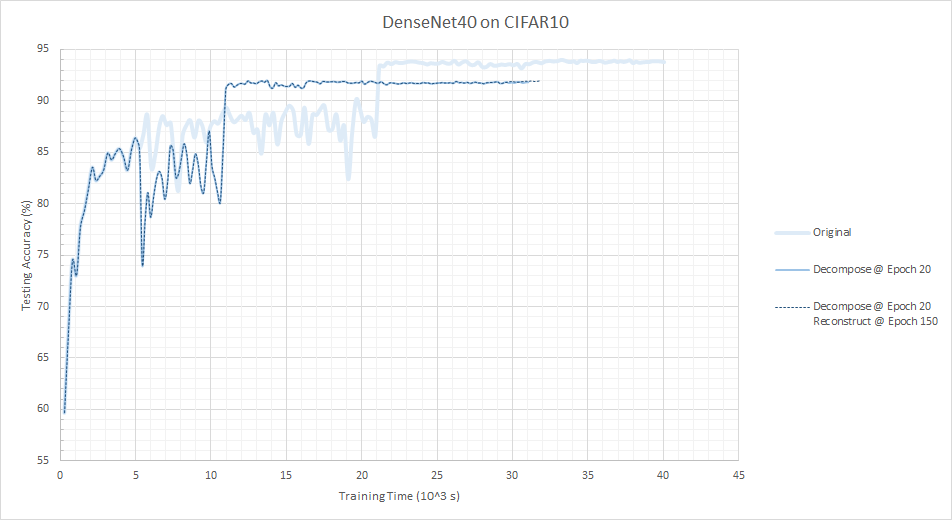}
	\caption{Training progress of DenseNet40 model on CIFAR10 dataset on NVIDIA Tesla P100 GPU for 160 epochs with batch size 128. } 
	\label{fig:cifar10-densenet40} 
\end{figure}

\begin{figure}[t]
	\centering
	\includegraphics[width=0.9\columnwidth]{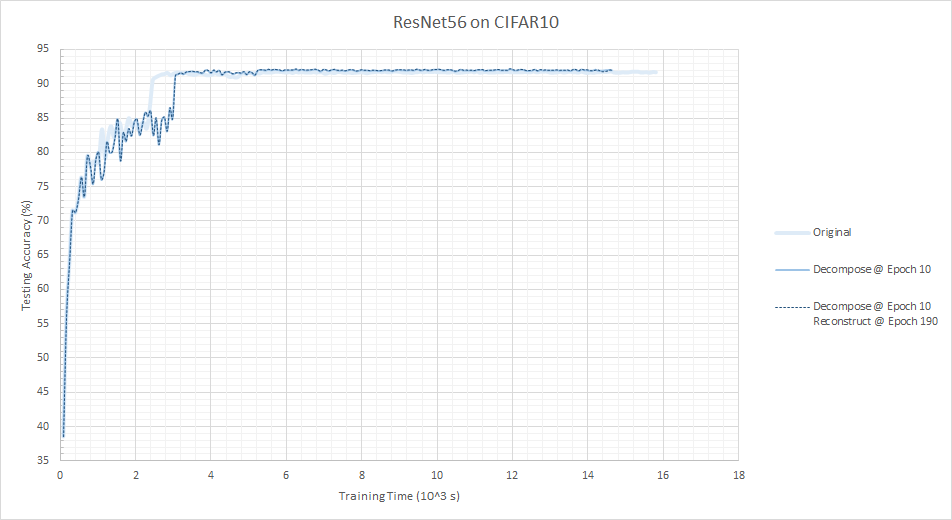}
	\caption{Training progress of ResNet56 model on CIFAR10 dataset on NVIDIA Tesla K40c GPU for 200 epochs with batch size 128.} 
	\label{fig:cifar10-resnet56}
\end{figure}

\begin{figure}
	\centering
	\begin{subfigure}[b]{0.9\columnwidth}
		\centering
		\includegraphics[width=\textwidth]{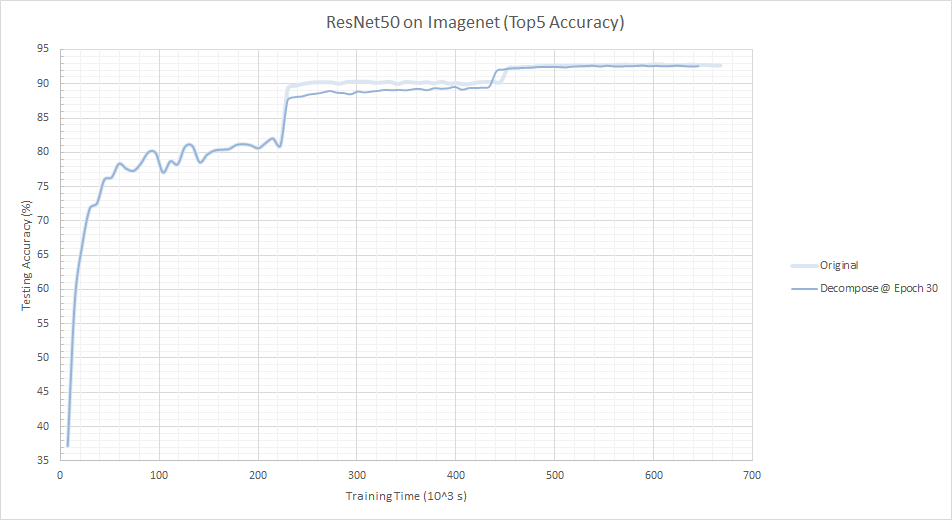}
		\caption{Top5 Accuracy}
		\label{fig:imagenet-resnet50-top5}
	\end{subfigure}
	\hfill
	\begin{subfigure}[b]{0.9\columnwidth}
		\centering
		\includegraphics[width=\textwidth]{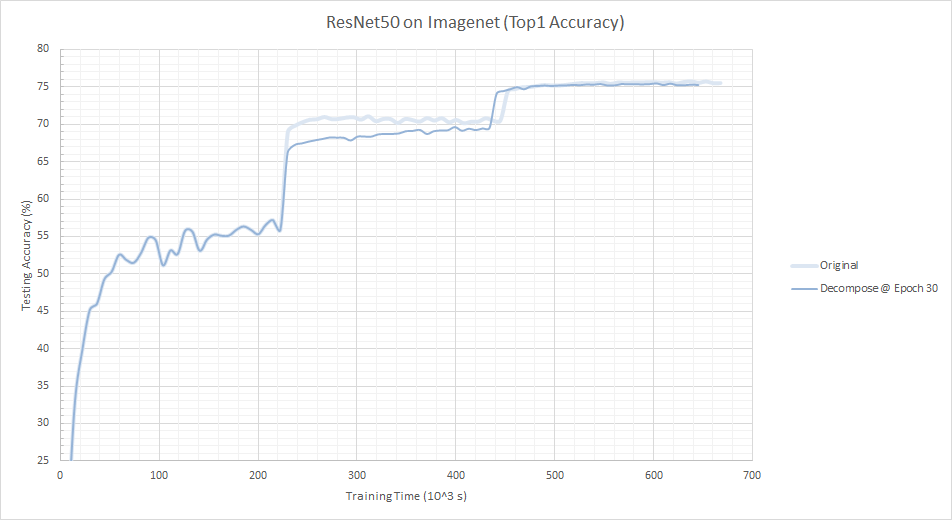}
		\caption{Top1 Accuracy}
		\label{fig:imagenet-resnet50-top1}
	\end{subfigure}
	\caption{Training progress of ResNet50 model on Imagenet dataset on NVIDIA Tesla V100 GPU for 90 epochs with batch size of 256.} 
	\label{fig:imagenet-resnet50}
\end{figure}

The results are shown in Tables \ref{table:cifar10-vgg19}, \ref{table:cifar10-densenet40}, \ref{table:cifar10-resnet56} and \ref{table:imagenet-resnet50} and Figures \ref{fig:cifar10-vgg19}, \ref{fig:cifar10-densenet40}, \ref{fig:cifar10-resnet56}, and \ref{fig:imagenet-resnet50}. In those tables, ``Dec.'' is abbreviation for decomposed, and ``Rec.'' is abbreviation for reconstructed.
\subsection{Results}
\begin{figure}[t]
	\centering
	\includegraphics[width=0.9\columnwidth]{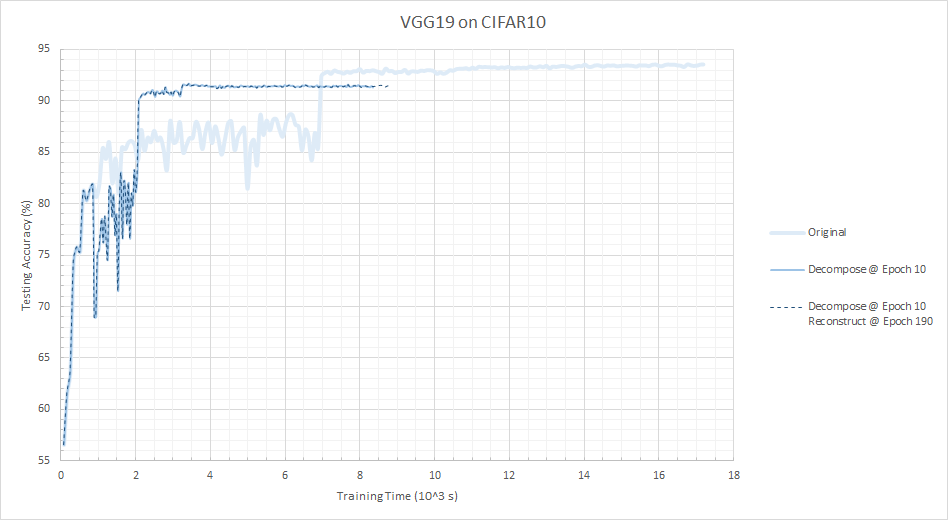}
	\caption{Training progress of VGG19 model on CIFAR10 dataset on NVIDIA Tesla K40c GPU for 200 epochs with batch size 128.} 
	\label{fig:cifar10-vgg19}
\end{figure}

For VGG19 on CIFAR10, we notice from Table \ref{table:cifar10-vgg19} that there was more than $2\times$ speedup in training time, a $20 \times$ compression of parameter size, but a drop of almost 2\% when decomposed from the 10th epoch. From Figure \ref{fig:cifar10-vgg19} we notice, a sudden drop in accuracy when decomposition happens, however that drop is compensated for after less than 1000 seconds. When decomposing from later epochs, there was a general trend of decreasing accuracy drop in return for a reduction in model size compression. It may seem that in later epochs, the EVBMF detects more noise in the weight values - as the weights try to fit the training data with higher accuracy and cover more corner cases - and hence selects a higher rank for decomposition. Surprisingly, the scenarios for reconstructing at the 190th epoch, and for decomposing after complete training, did not result in higher best accuracy than decomposing at the 10th epoch without reconstruction.

For DenseNet40 on CIFAR10, we notice a similar drop in accuracy as in VGG19, but less training speedup, despite more than $3 \times$ reduction in the number of parameters. This is expected from the compression and speedup ratios expressed in Equations \ref{eqn:compression} and \ref{eqn:speedup}. The results for DenseNet40 also show that both accuracy and model compression are higher for decomposing during training than decomposing after training.

On the other hand, as shown in Table \ref{table:cifar10-resnet56}, decomposing ResNet56 resulted in an increase accuracy, but less than 10\% reduction in training time. This can be interpreted that the EVBMF algorithm selected high ranks for most layers in the ResNet56 model. This is a point that should be further researched in the future for analysis.

For Imagenet dataset, the drop in accuracy was less than 0.2\% for ResNet50 but the reduction in training time was negligible. Furthermore, decomposing at the 30th epoch resulted in better accuracy than decomposing after complete training.

\begin{table}
	\caption{Performance and size of VGG19 with different scenarios of training on NVIDIA Tesla K40c GPU on CIFAR10 dataset. The epoch at which decomposition or reconstruction happens is mentioned. The total number of epochs for all scenarios is 200, except for the last case where decomposition happens after the 200th epoch, and fine tuned for another 40 epochs. }
	\label{table:cifar10-vgg19}
	\centering
	\begin{tabular}{lllll}
		\toprule
		Model & \multicolumn{2}{c}{Accuracy} & Params & \shortstack[h]{Training \\ Time} \\ 
		\cmidrule(r){2-3}
		& Best & Final & &  \\
		\midrule
		Original & 93.55\% & 93.56\% & $20\times 10^6$ & 4.77 hr \\
		\midrule
		Dec. @ 10 & 91.69\% & 91.39\% & $749\times 10^3$ & 2.32 hr \\
		Dec. @ 20 & 92.10\% & 91.89\% & $1.33\times 10^6$ & 2.75 hr \\
		Dec. @ 30 & 91.85\% & 91.78\% & $1.69\times 10^6$ & 2.95 hr \\
		Dec. @ 40 & 92.57\% & 92.43\% & $1.75\times 10^6$ & 3.04 hr \\
		Dec. @ 50 & 92.51\% & 92.49\% & $1.73\times 10^6$ & 2.41 hr \\
		\midrule
		\blap{Dec. @ 10 \\ Rec. @ 190} & 91.69\% & 91.53\% & $749\times 10^3$ & 2.45 hr \\
		\midrule
		\blap{Original \\ then Dec.} & 91.52\% & 91.36\% & $1.33\times 10^6$ & \blap{4.77 hr \\ + 0.49 hr} \\  
		\bottomrule
	\end{tabular}
\end{table}

\begin{table}
	\caption{Performance and size of DenseNet40 with different scenarios of training and decomposition on NVIDIA Tesla P100 GPU on CIFAR10 dataset. The total number of epochs for all scenarios is 160, except for the last case where decomposition happens after the 160th epoch, and fine tuned for another 40 epochs.}
	\label{table:cifar10-densenet40}
	\centering
	\begin{tabular}{lllll}
		\toprule
		Model & \multicolumn{2}{c}{Accuracy} & Params & \shortstack[h]{Training \\ Time} \\ 
		\cmidrule(r){2-3}
		& Best & Final & &  \\
		\midrule
		Original & 94.00\% & 93.78\% & $1.06\times 10^6$ & 11.13 hr \\
		\midrule
		Dec. @ 20 & 92.00\% & 91.82\% & $270\times 10^3$ & 8.62 hr \\
		\midrule
		\blap{Dec. @ 20 \\ Rec. @ 150} & 92.00\% & 91.96\% & $1.06\times 10^6$ & 8.83 hr \\
		\midrule
		\blap{Original \\ then Dec.} & 91.49\% & 91.46\% & $441\times 10^3$  & \blap{11.13 hr \\ + 2.17 hr} \\  
		\bottomrule
	\end{tabular}
\end{table}

\begin{table}
	\caption{Performance and size of ResNet56 with different scenarios of training on NVIDIA Tesla K40c GPU on CIFAR10 dataset. The epoch at which decomposition or reconstruction happens is mentioned. The total number of epochs for all scenarios is 200, except for the last case where decomposition happens after the 200th epoch, and fine tuned for another 40 epochs. }
	\label{table:cifar10-resnet56}
	\centering
	\begin{tabular}{lllll}
		\toprule
		Model & \multicolumn{2}{c}{Accuracy} & Params & \shortstack[h]{Training \\ Time} \\ 
		\cmidrule(r){2-3}
		& Best & Final & &  \\
		\midrule
		Original & 91.83\% & 91.69\% & $853\times 10^3$ & 4.39 hr \\
		\midrule
		Dec. @ 10 & 92.16\% & 91.97\% & $508\times 10^3$ & 4.06 hr \\
		Dec. @ 20 & 92.27\% & 92.07\% & $520\times 10^3$ & 4.10 hr \\
		Dec. @ 30 & 91.66\% & 91.51\% & $556\times 10^3$ & 4.15 hr \\
		Dec. @ 40 & 91.67\% & 91.50\% & $550\times 10^3$ & 4.14 hr \\
		Dec. @ 50 & 91.65\% & 91.15\% & $550\times 10^6$ & 4.15 hr \\
		\midrule
		\blap{Dec. @ 10 \\ Rec. @ 190} & 92.16\% & 91.92\% & $853\times 10^3$ & 4.07 hr \\
		\midrule
		\blap{Original \\ then Dec.} & 92.32\% & 92.22\% & $547\times 10^3$ & \blap{4.39 hr \\ + 0.85 hr} \\  
		\bottomrule
	\end{tabular}
\end{table}

\begin{table}
	\caption{Performance and size of ResNet50 with different scenarios of training on NVIDIA Tesla K40c GPU on Imagenet dataset. The epoch at which decomposition or reconstruction happens is mentioned. The total number of epochs for all scenarios is 90, except for the last case where decomposition happens after the 90th epoch, and fine tuned for another 20 epochs.}
	\label{table:imagenet-resnet50}
	\centering
	\begin{tabular}{lllll}
		\toprule
		Model & \multicolumn{2}{c}{Best Accuracy} & Params & \shortstack[h]{Training \\ Time} \\ 
		\cmidrule(r){2-3}
		& Top1 & Top5 & &  \\
		\midrule
		Original & 75.65\% & 92.85\% & $25.6\times 10^6$ & 185.4 hr \\
		\midrule
		Dec. @ 30 & 75.34\% & 92.68\% & $17.6\times 10^6$ & 179.2 hr \\
		\midrule
		\blap{Original \\ then Dec.} & 69.26\% & 89.38\% & $8.2\times 10^6$ & \blap{185.4 hr \\ + 20.56 hr} \\  
		\bottomrule
	\end{tabular}
\end{table}


\section{Conclusion and Future Work}
In this paper we have shown that to compress a model using tensor decomposition, we do not have to wait till training ends. We have shown the decomposing at the 10th or 20th epoch of training, results in accuracy close to - and sometimes higher than - that of the original model trained till the end.

We have also shown that in all of the cases on CIFAR10 dataset, the size of a model decomposed after 10 or 20 epochs of training is smaller than that of the model decomposed after complete training.
Moreover, we have shown - for CIFAR10 - that training a decomposed model for VGG and DenseNet architectures results in considerable faster training time: more than $2\times$ for VGG19 and $1.3\times$ for DenseNet40. However, the speedup obtained for ResNet architecture was negligible.

For future work, there is a need to explore ways to reduce the accuracy drop in accuracy for our ``decomposition-in-training'' approach for some models, and to increase the training speedup for other architectures especially ResNet.  

\section{Acknowledgments}
We would like to thank Jacob Gildenblat \cite{JGildenblat2018} and Ruihang Du \cite{RDu2018} for providing open-sourced code for Tucker decomposition using PyTorch. We also thank Yerlan Idelbayev for providing open-sourced code and model files to reproduce the accuracies of the original ResNet \cite{pytorch_cifar10}, DenseNet, and VGG papers \cite{EMingjie2018} results on CIFAR10. 

\bibliographystyle{aaai}
\bibliography{paper}

\end{document}